\PassOptionsToPackage{table}{xcolor}
\documentclass[sigconf]{acmart}

\usepackage{amsmath,amsfonts}

\usepackage{algpseudocode}
\usepackage{graphicx}
\usepackage{algorithm}
\usepackage{algpseudocode}
\usepackage{multirow}
\usepackage{subcaption}
\usepackage{booktabs}
\usepackage{enumitem}
\usepackage{balance}

\copyrightyear{2026}
\acmYear{2026}
\setcopyright{usgovmixed}
\acmConference[HPDC '26]{The 35th International Symposium on High-Performance Parallel and Distributed Computing}{July 13--16, 2026}{Cleveland, OH, USA}
\acmBooktitle{The 35th International Symposium on High-Performance Parallel and Distributed Computing (HPDC '26), July 13--16, 2026, Cleveland, OH, USA}
\acmDOI{10.1145/3806645.3816237}
\acmISBN{979-8-4007-2640-8/2026/07}

\begin{document}

\title{Beyond Fixed Budgets: Characterizing the Inelasticity and Limitations of Tree-of-Thought Reasoning Strategies}

\settopmatter{authorsperrow=4}
\author{Atkia Mahila}
\affiliation{
  \institution{Rochester Institute of Technology}
  \city{Rochester}
  \country{USA}}
\email{am9183@rit.edu}

\author{Avinash Maurya}
\affiliation{
   \institution{Argonne National Laboratory}
   \city{Lemont}
   \country{USA}}
\email{amaurya@anl.gov}

\author{M. Mustafa Rafique}
\affiliation{
  \institution{Rochester Institute of Technology}
  \city{Rochester}
   \country{USA}}
\email{mrafique@cs.rit.edu}

\author{Bogdan Nicolae}
\affiliation{
   \institution{Argonne National Laboratory}
    \city{Lemont}
   \country{USA}}
\email{bnicolae@anl.gov}

\renewcommand{\shortauthors}{Mahila et al.}

\begin{abstract}
Tree of Thought (ToT) search has become a promising direction for improving the reasoning
capabilities of large language models, but deploying these methods in practice raises a
question that has received little systematic attention: how do different search strategies
behave under varying compute budgets, model sizes, and problem difficulties? In this work,
we evaluate two representative ToT methods; DPTS, a Monte Carlo tree search based approach,
and SSDP, a semantic deduplication based approach, across two mathematical reasoning
benchmarks (Math500 and GSM8K), two model scales (Llama-3B and Llama-8B), and four token
budgets (3k--10k). Our analysis reveals that the two methods exhibit limitations that pull in
opposite directions. DPTS suffers from a cold-start bottleneck at low budgets: it requires sufficient
exploration before its value estimates become reliable, making it a poor fit for
resource-constrained settings despite strong scaling behavior at higher budgets. SSDP,
on the other hand, reaches candidate solutions efficiently but is prone to frontier
depletion; its aggressive node merging permanently discards unexplored paths, leaving
it unable to improve regardless of how much budget remains. Together,
these findings suggest that neither a fixed exploration strategy nor a fixed pruning
strategy is sufficient across compute continuum. We argue that effective search for
scientific reasoning agents requires strategies that can adapt their behavior based on
search progress and available resources.
\end{abstract}




\maketitle

\section{Introduction}
\label{sec:intro}
Large language models (LLMs) and transformer-based reasoning agents are rapidly becoming integral components of modern scientific workflows, supporting tasks that range from automated literature analysis and hypothesis generation to symbolic mathematics, code synthesis, and physics-aware decision making~\cite{touvron2023llama, hoffmann2022training, wei2022chain}. A central enabler of this trend is \emph{inference-time scaling}: rather than relying on a single autoregressive pass or a chain of inferences ~\cite{wei2022chain}, modern reasoning systems explore, evaluate, and refine multiple alternative reasoning trajectories~\cite{yao2023tree, kaplan2020scaling}. Tree-of-Thought (ToT) search has emerged as the dominant abstraction for this paradigm, formulating reasoning as an explicit tree traversal where nodes encode partial reasoning states and edges correspond to model-generated expansions. However, with increasing token cost, maximizing token budgets allocated for ToT reasoning becomes a paramount problem. In this case, ToT is subject to an important trade-off: how many alterative branches to explore vs. how deep in each branch, how far branches should be apart to mitigate the risk of being trapped in a local optimum without wasting tokens on irrelevant directions, etc. Such trade-offs are especially relevant
for flexible scientific computing ecosystems, where multi-tenant clusters, elastic cloud bursts and shared GPU pools impose dynamic and heterogeneous compute constraints, resulting in highly variable per-query token budget that such reasoning agents can afford~\cite{characterizingInferencesIPDPS2025,arif2026understanding}. Thus,
gaining insight into the behavior of ToT reasoning strategies is a crucial intermediate
step in the design of adaptive exploration strategies, whose scalability and cost-effectiveness directly determine the end-to-end utility of modern AI scientific workflows.

\paragraph{\bf Limitations of State-of-Art.}
Despite the rapid proliferation of ToT variants, the community lacks a systematic understanding of \emph{how} representative reasoning strategies actually traverse the tree of thought under realistic, fluctuating resource constraints. Existing evaluations predominantly report aggregate accuracy at fixed, often generous, compute budgets~\cite{xie2024monte, kim2025chopping, li2025mits}, leaving the internal search dynamics (such as  when candidate solutions first emerge, why and where the search terminates, and how additional tokens are converted into deeper or broader exploration) largely opaque.
Specifically, Monte Carlo Tree Search (MCTS) approaches~\cite{xie2024monte, ding2025dynamic, gao2024interpretable} rely on iterative rollouts and value backpropagation, a paradigm that is statistically powerful but subject to an expensive warm-up phase whose cost has rarely been quantified in budget-constrained regimes. Semantic similarity-based pruning approaches~\cite{kim2025chopping, wang2025don} aggressively merge redundant reasoning branches to amortize token cost, but the long-term impact of this merging on search frontier diversity remains poorly characterized. Critically, both families embed \emph{fixed} algorithmic policies (rollout counts, similarity thresholds, gating rules) that are decided at design time, not at inference time. On flexible infrastructures where compute constraints are dynamically fluctuating, this rigidity translates into a significant disadvantage: methods that look strong in one budget regime can collapse in another, and additional compute often fails to translate into proportional accuracy gains~\cite{cao2024flexible}. This is the gap our work fills.

\paragraph{\bf Key Insights and Contributions.}
This paper argues that delivering flexible, resource-adaptive scientific reasoning first requires opening the black box of ToT search and characterizing how representative strategies explore the tree of thought as a function of compute budget, model scale, and problem difficulty. Through controlled experiments on two representative ToT methods, two mathematical reasoning benchmarks, two Llama model scales, and four token budgets, we identify two limitations that pull in opposite directions, so any fixed policy that avoids one runs into the other, and we use this evidence to articulate concrete requirements for the next generation of flexible ToT engines. We summarize our contributions as follows:

\begin{enumerate}[topsep=0pt,itemsep=0pt,leftmargin=12pt]
\item \textbf{Search-dynamics characterization across the compute continuum:} We conduct a systematic side-by-side empirical study of how an MCTS-based strategy (DPTS~\cite{ding2025dynamic}) and a semantic-pruning strategy (SSDP~\cite{kim2025chopping}) traverse the tree of thought under varying compute budgets. Beyond aggregate accuracy, we instrument candidate reachability, termination causes, and per-budget token utilization to reveal \emph{how} each method actually spends the compute it is given (\S~\ref{sec:methods},~\ref{sec:exp}).

\item \textbf{Two opposing limitations:} We show that DPTS suffers a structural \emph{cold-start bottleneck}, failing to produce any candidate solution on up to 74\% of Math500 problems at a 3k budget because its value estimates remain noisy before rollout statistics stabilize. Conversely, SSDP exhibits \emph{frontier depletion}: its aggressive merging permanently discards unexplored branches, causing accuracy and token usage to plateau regardless of how much additional budget is granted (\S~\ref{sec:exp}).

\item \textbf{Diagnosis of budget inelasticity:} We quantify the gap between budget \emph{allocation} and budget \emph{utilization}, showing that SSDP's average token usage stays nearly flat (around 2k) whether 3k or 10k tokens are available, while DPTS only converts additional tokens into accuracy gains after its warm-up cost has been amortized. This exposes a fundamental misalignment between fixed ToT policies and the elastic, heterogeneous resource envelopes typical of flexible scientific infrastructures (\S~\ref{sec:exp}).

\item \textbf{Implications for flexible ToT design:} We translate these findings into concrete requirements for adaptive reasoning strategies that jointly balance accuracy, token budget, and scalability across heterogeneous infrastructures, by shifting between exploration and pruning-dominated regimes based on search progress and remaining budget rather than committing to a single policy at design time (\S~\ref{sec:conclusions}).
\end{enumerate}

\section{Related Work}
\label{sec:related}

\paragraph{\bf From Chain-of-Thought to Tree-of-Thought.}
Inference-time scaling has rapidly displaced single-pass decoding for complex reasoning. Chain-of-Thought (CoT) prompting~\cite{wei2022chain} showed that eliciting intermediate reasoning steps substantially improves arithmetic and symbolic accuracy, and self-consistency decoding~\cite{wang2022self} extended this by sampling and aggregating multiple flat reasoning paths. These approaches lack explicit structure over the reasoning process. Tree-of-Thought (ToT)~\cite{yao2023tree} reframes inference as search over latent reasoning trajectories, where nodes encode partial reasoning states and edges correspond to model-generated expansions. This abstraction enables the application of classical search strategies such as breadth-first and depth-first search~\cite{yao2023tree}, heuristic-guided beam search~\cite{wang2025bpp}, and LLM-guided exploration~\cite{herr2025llm}, but it also raises the central question of \emph{which} search policy to apply under realistic compute constraints.

\paragraph{\bf MCTS-based ToT}
A prominent line of work adapts Monte Carlo Tree Search~\cite{swiechowski2023monte, browne2012survey} to reasoning, balancing exploration and exploitation through iterative rollouts and value estimation. Several methods treat the LLM as a world model and combine likelihood, self-evaluation, environment feedback, or partial search into custom reward functions~\cite{feng2023alphazero, lin2026cmcts, ding2025dynamic}, while SC-MCTS*~\cite{gao2024interpretable} adds a contrastive reward model, speculative decoding for speed, and improved UCT selection and backpropagation. Other MCTS variants are used to collect training data or to drive RL-style training with specialized rewards and selection strategies~\cite{xie2024monte, li2025rethinkmcts, yao2024mulberry, xiong2025mcts}. A property shared by all of these approaches is their reliance on a non-trivial volume of initial samples~\cite{xie2024monte} before value estimates become informative; under tight token budgets, this manifests as a startup bottleneck where the search may exhaust its budget before producing any viable candidate.

\paragraph{\bf Pruning-based ToT}
A complementary line of work attacks the cost of exhaustive exploration by pruning the search space. Early approaches use heuristic filtering~\cite{wang2025litesearch} or beam-search-style truncation~\cite{jiang2024enhancing}, while more recent methods exploit semantic signals: FETCH~\cite{wang2025don} clusters semantically equivalent states via embedding similarity and stabilizes verifiers to reduce under-exploration; SSDP~\cite{kim2025chopping} dynamically merges similar reasoning paths online during ToT; and MITS~\cite{li2025mits} scores steps via pointwise mutual information and uses beam search with entropy-based dynamic sampling to focus on uncertain, informative steps. While these methods substantially reduce token cost, aggressive pruning can collapse the diversity of the reasoning space~\cite{kim2025chopping, miyamoto2026aligning}, removing alternative paths needed to recover from early errors and ultimately leading to frontier depletion.

\paragraph{\bf Adaptive search and the gap addressed in this work.}
The broader machine learning community has long recognized that fixed algorithmic choices fail under distribution shift or varying problem conditions~\cite{zhang2021adaptive}; adaptive online algorithms that respond to problem properties, such as those that adjust learning rates based on gradient variance~\cite{duchi2011adaptive}, have consistently proven more robust. ToT search exhibits the same need for adaptivity: search controllers should dynamically adjust their policies based on the available budget, model scale, and current search phase to ensure compute is neither wasted nor prematurely restricted. While individual studies have observed startup or pruning issues in isolation, there is limited work that systematically evaluates these limitations side-by-side across methods, budgets, and model scales. In this paper, we focus on two representative paradigms, MCTS-based search (DPTS~\cite{ding2025dynamic}) and semantic similarity-based pruning (SSDP~\cite{kim2025chopping}), and conduct a controlled empirical analysis that isolates their fundamental bottlenecks and motivates the need for flexible, resource-adaptive reasoning strategies.

\section{Methodology}
\label{sec:methods}

Our contribution focuses on the instrumentation and side-by-side study of two representative existing strategies ToT strategies, namely DPTS~\cite{ding2025dynamic} (as a MCTS representative), and SSDP~\cite{kim2025chopping} (as a semantic-pruning representative). We selected these two methods because they sit at opposite ends of the design spectrum identified in \S~\ref{sec:related}, i.e., rollout-driven exploration vs. semantic-similarity-driven pruning, and because both are recent, peer-reviewed, and ship reproducible reference implementations. We re-implemented both methods within a common evaluation harness, used the original hyperparameter settings reported by their authors to avoid confounding our findings with tuning artifacts, and varied only the per-problem token budget across experiments.

\subsection{Experimental Setup}
\label{sec:setup}

\paragraph{\bf Testbed.}
We deployed our evaluation on the ALCF Polaris supercomputer \cite{alcf_polaris}. Polaris consists of 560 compute nodes, each equipped with a single 2.8~GHz 32-core AMD EPYC Milan CPU, 512~GB of DDR4 memory, and 4$\times$NVIDIA A100 GPUs. We leverage up to 64 compute nodes with shared access to a 10~TB Lustre PFS.

\paragraph{\bf Datasets.}
To evaluate the elasticity and robustness of these search methods, we utilize datasets representing a broad scale of logical complexity and reward sparsity: (1) \emph{GSM8K:} includes 1,319 problems requiring basic multi-step arithmetic \cite{cobbe2021training}. It serves as a baseline for high-reachability search scenarios where limitations are typically localized to calculation errors rather than search collapse; (2) \emph{Math500:} a comprehensive set of 500 problems spanning diverse difficulty levels and mathematical domains \cite{hendrycks2021measuring}. Unlike simpler benchmarks, Math500 contains problems with varying reasoning depths, where higher difficulty instances require significantly deeper trajectories and more precise step by step verification to reach a correct final state.

\paragraph{\bf Models and Constraints.}
The experiments are conducted using \emph{Llama-3.2-3B} and \emph{Llama-3.1-8B} to observe the interaction between model scale and search dynamics. To rigorously test resource adaptation, we enforce per-problem generation budgets of \textbf{3k, 5k, 8k, and 10k output tokens}. In practice, the budget is enforced after each batched expansion step. Since methods may issue multiple generation requests within one expansion round, the total generated-token count can slightly exceed the nominal budget before the stopping condition is checked.

\subsection{Compared Approaches}
\label{sec:compared}
\paragraph{\bf DPTS (Dynamic Parallel Tree Search).}
DPTS is governed by a Search and Transition module that manages compute allocation between exploration and exploitation. Rather than a rigid expansion policy, DPTS uses a dynamic stopping and gating policy: in this study, we configure a fixed tree width of 4 and a maximum tree depth of 16 (matching the original authors' setting), and a quality-gating hyperparameter $t^* = 5$ controls when the algorithm transitions from free exploration to selective expansion. During the initial phase, the search freely explores the reasoning space; once $t^*$ terminal solutions are found, a strict best-solution-aware stopping rule applies and a new rollout is pursued only if the selected node's reward strictly exceeds the best known terminal reward. Like all MCTS-style methods, DPTS depends on an initial sampling volume to make its value estimates informative: until enough rollouts have been completed at relevant depths, node values remain noisy and the policy effectively explores at random. This is the structural origin of the \emph{cold-start bottleneck} we measure in \S~\ref{sec:exp}, and it becomes more severe on problem distributions that require deep reasoning (e.g., Math500, with 4--16 reasoning steps) than on shallower ones (e.g., GSM8K, with 3--8 steps), because the number of rollouts needed to stabilize the statistics grows with tree depth at a fixed branching factor.

\paragraph{\bf SSDP (Semantic Similarity Dynamic Pruning).}
SSDP is an efficiency-oriented extension of DPTS that mitigates branch explosion by exploiting semantic redundancy. At the search frontier, every newly generated reasoning node is mapped to a high-dimensional embedding, and its cosine similarity to existing siblings is used to detect overlapping reasoning logic. A fixed similarity threshold $\tau = 0.75$ governs the pruning decision: if two siblings exceed this threshold, they are merged and the unexplored children of the merged-away node are discarded permanently.
The irrevocable nature of this merging is the structural origin of the \emph{frontier depletion} we measure in \S~\ref{sec:exp}: if $\tau$ is too permissive for the embedding space at hand, semantically distinct reasoning paths are collapsed together, the frontier drains before a viable solution is found, and no amount of additional budget can resurrect the discarded branches. Because the geometry of the embedding space is itself a property of the underlying model, the same fixed $\tau$ can behave very differently across model scales: lower-capacity models tend to produce less spread-out embeddings~\cite{yuan2026ecr, vseverdija2023compressing}, which compresses genuinely different reasoning paths into nearby regions and effectively makes SSDP markedly more aggressive on smaller models than on larger ones at the same threshold.

\subsection{Instrumentation}
\label{sec:instrumentation}

To surface the internal search dynamics that aggregate accuracy alone cannot reveal, we instrumented both methods with a shared set of probes that report, for every problem instance:

\begin{itemize}[topsep=0pt,itemsep=0pt,leftmargin=12pt]
\item \textbf{Final accuracy}: whether the search produced a correct final answer. This is the standard metric reported in prior ToT work and serves as our outermost performance signal.
\item \textbf{Candidate reachability}: whether the search produced \emph{any} terminal candidate solution within the allocated budget, regardless of correctness. Reachability isolates whether the algorithm can complete a trajectory at all, decoupled from solution quality.
\item \textbf{Token utilization}: the average number of tokens actually generated per problem, contrasted against the allocated per-problem token budget. The gap between allocation and utilization quantifies how elastically a method consumes additional compute when it is offered.
\item \textbf{Termination cause}: the reason the search stopped on a given instance, attributed to one of three categories: \emph{budget exhaustion} (the token cap was reached), \emph{early stopping} (the method's internal gating decided to stop), or \emph{frontier depletion} (no expandable nodes remained in the tree). Termination causes reveal whether the search is bound by compute or by its own design choices.
\end{itemize}

\paragraph{\bf Relevance of metrics.} These probes are deliberately chosen to operationalize the insights highlighted as focus areas in \S~\ref{sec:intro}. First, the contrast between \emph{reachability} and \emph{accuracy} isolates the \emph{cold-start} regime: a method whose reachability is very low at small budgets cannot produce candidates and therefore cannot be evaluated on accuracy at all, making cold-start visible long before it manifests as an accuracy collapse. Second, the \emph{termination-cause} breakdown distinguishes budget-bound from frontier-bound stopping: a method that terminates predominantly through frontier depletion has nothing left to expand and cannot benefit from additional tokens, regardless of how generous the budget is, which is the lens through which the SSDP plateau becomes visible. Third, \emph{token utilization} measured against the allocated budget directly quantifies budget inelasticity: a method whose generated-token count plateaus far below the cap is structurally unable to convert additional compute into additional search effort. Together, these probes expose how each strategy actually traverses the tree of thought rather than merely whether it eventually succeeds, which is precisely the visibility the community currently lacks (\S~\ref{sec:intro}).

\begin{table}[h]
\centering
\caption{Search reachability (\% of problems generating at least $1$
candidate answer) across token budgets.Higher is better. {\color{red} Red} indicates
low-reachability settings.}
\label{tab:reachability_cliff}
\resizebox{\columnwidth}{!}{%
\begin{tabular}{lllcccc}
\toprule
\textbf{Dataset} & \textbf{Model} & \textbf{Method} & \textbf{3k Budget} & \textbf{5k Budget} & \textbf{8k Budget} & \textbf{10k Budget} \\ \midrule
\multirow{4}{*}{Math500} & \multirow{2}{*}{Llama-3B} & DPTS & \color{red}\textbf{29.4\%} & 52.2\% & 66.2\% & 71.2\% \\
 &  & SSDP & 68.8\% & 79.4\% & 84.0\% & 81.8\% \\ \cmidrule{2-7}
 & \multirow{2}{*}{Llama-8B} & DPTS & \color{red}\textbf{26.0\%} & 46.6\% & 62.2\% & 68.8\% \\
 &  & SSDP & 67.4\% & 76.2\% & 81.8\% & 82.0\% \\ \midrule
\multirow{4}{*}{GSM8K} & \multirow{2}{*}{Llama-3B} & DPTS & 50.0\% & 62.9\% & 67.6\% & 69.7\% \\
 &  & SSDP & 42.7\% & 42.4\% & 43.4\% & 41.7\% \\ \cmidrule{2-7}
 & \multirow{2}{*}{Llama-8B} & DPTS & 67.4\% & 92.0\% & 98.4\% & 99.2\% \\
 &  & SSDP & 94.2\% & 93.6\% & 93.6\% & 93.6\% \\ \bottomrule
\end{tabular}%
}
\end{table}

\section{Experimental Evaluation}
\label{sec:exp}

\subsection{Monte Carlo Cold-Start Bottleneck in DPTS}

Table~\ref{tab:reachability_cliff} measures cold-start regime described in \S~\ref{sec:compared} quantitatively. On Math500 with Llama-3B, DPTS produces a candidate answer on only 29.4\% of problems at a 3k budget, which is fewer than one in three. The Llama-8B result is even harsher at 26.0\% reachability, meaning that for up to 74\% of problems no complete reasoning trajectory is generated at all. As the budget grows, reachability climbs to 71.2\% and 68.8\% respectively at 10k, and Figure~\ref{fig:accuracy_tokens} shows accuracy tracking the same trajectory (e.g., 23.4\% to 45.5\% on Math500 with Llama-3B). The asymmetry between Math500 and GSM8K confirms the depth dependence noted in \S~\ref{sec:compared}: on the shallower GSM8K, DPTS already reaches 50.0\% (Llama-3B) and 67.4\% (Llama-8B) reachability at the same 3k budget where Math500 collapses.

These observations are of high consequence with respect to flexible infrastructures.
In resource-constrained settings, which the 3k budget point is representative of, DPTS's reachability collapses, meaning that for a large fraction of problems no answer is produced at all. This automatically disqualifies it as a viable strategy for small budgets, despite its strong scaling at higher budgets.

\subsection{Frontier Depletion: How SSDP's Efficiency Becomes Its Limitation}

\begin{figure}
    \centering
    \includegraphics[width=\linewidth]{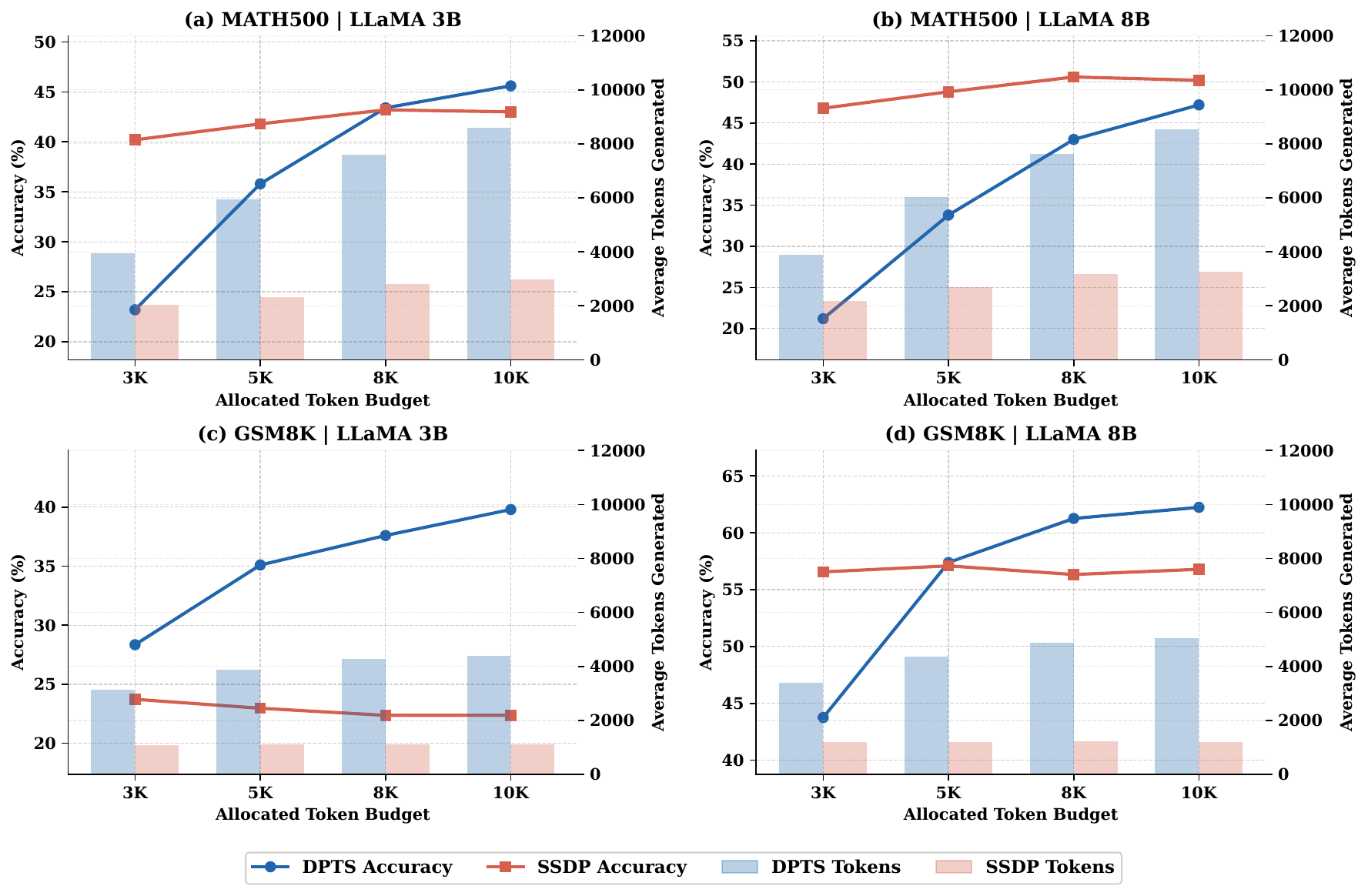}
    \caption{Accuracy (\%) and average tokens generated for DPTS and SSDP across allocated token budgets on Math500 and GSM8K with Llama-3B and Llama-8B. Left axis shows accuracy (lines); right axis shows average tokens generated (bars). }
    \label{fig:accuracy_tokens}
    \vspace{-10pt}
\end{figure}

SSDP's pruning strategy delivers on its primary promise of fast first-solution reachability. At a 3k budget on Math500, SSDP reaches a candidate on 68.8\% of problems with Llama-3B and 67.4\% with Llama-8B (Table~\ref{tab:reachability_cliff}), which is more than double the DPTS reachability in the same regime. On Math500 with Llama-8B it opens at 47.8\% accuracy at 3k, a lead it holds across most of the budget range (Figure~\ref{fig:accuracy_tokens}).

The frontier-depletion risk discussed in \S~\ref{sec:compared} materializes most visibly on GSM8K with Llama-3B, where SSDP collapses to 23.73\% accuracy, well below its own Llama-8B result of 56.56\% on the same dataset, and even below the DPTS baseline of 28.35\% on the same setup. Figure~\ref{fig:stopping_reasons} (bottom-left) makes the cause concrete: the frontier-depletion termination count for SSDP is 1,308 out of roughly 1,319 problems at 3k, and barely moves as the budget grows. The model-scale asymmetry is consistent with the embedding-geometry argument given in \S~\ref{sec:compared}: at the same $\tau$, the smaller model's tighter embedding space causes the merging step to over-collapse the frontier (e.g., two genuinely different paths might land at similarity $\approx 0.88$ under Llama-3B but $\approx 0.65$ under Llama-8B, only one of which crosses $\tau = 0.75$).

The consequence is the mirror image of DPTS: SSDP can start, but it cannot continue. Once the frontier is exhausted, doubling or tripling the budget yields no improvement at all, a structural ceiling that compute alone cannot break through. For flexible infrastructures where additional budget might be opportunistically available (a bursting cloud allocation, a temporarily idle GPU partition), SSDP simply cannot exploit it.

\subsection{Budget Utilization: Scaling Versus Saturation}

The budget-utilization view (Figure~\ref{fig:accuracy_tokens}) makes the two limitations directly comparable on the same axis. DPTS's average token usage grows in proportion with the allocated budget (from around 4k at the 3k budget level to 8--9k at 10k) and accuracy tracks that growth. SSDP's token usage, by contrast, stays nearly flat at roughly 2k across every budget level, and on GSM8K with Llama-3B it barely registers on the axis while accuracy stagnates around 23--25\%.

\begin{figure}[h]
    \centering
    \includegraphics[width=\linewidth]{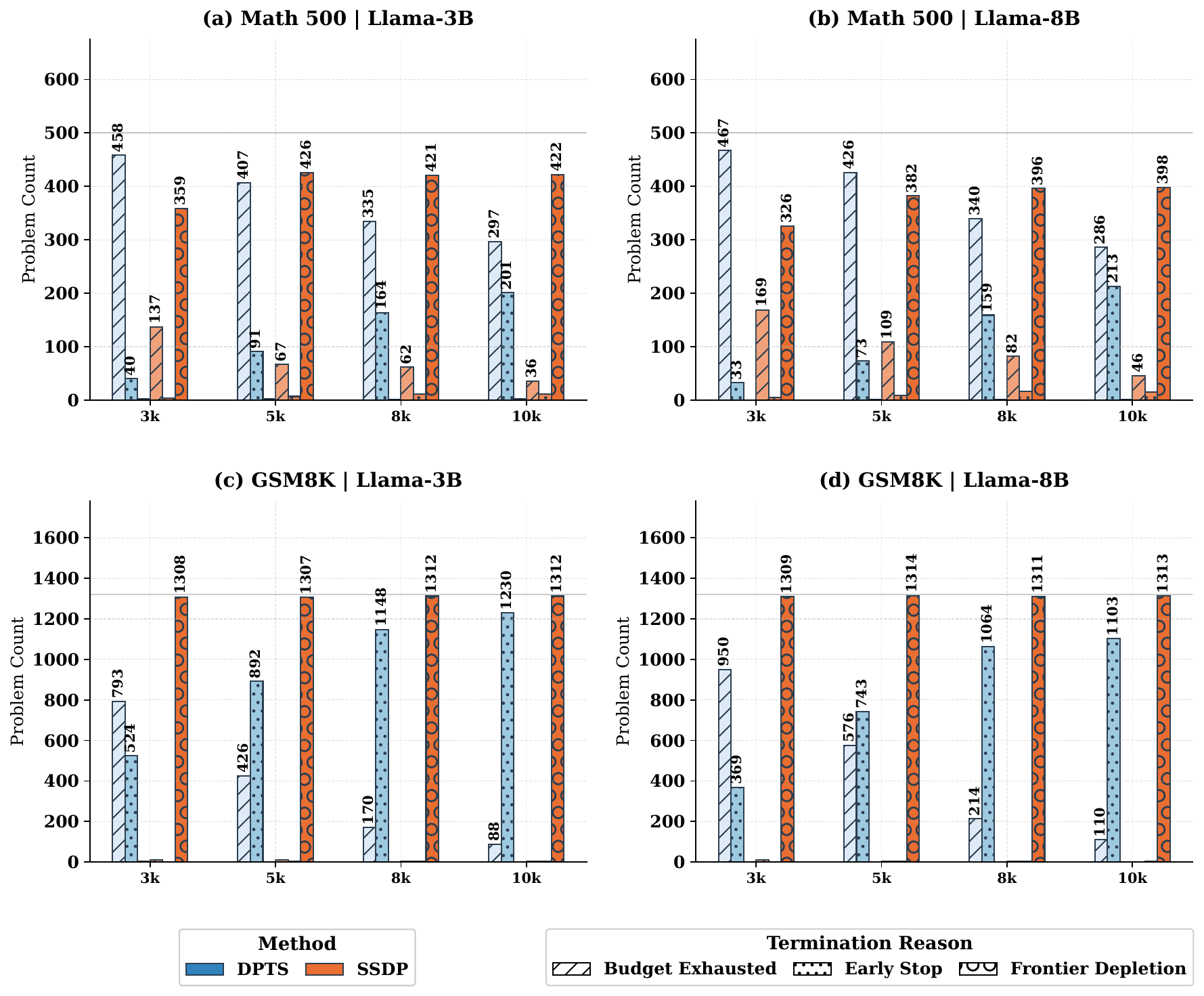}
    \caption{Stopping reason distribution across budgets for DPTS (blue) and SSDP (orange) on Math500 and GSM8K with Llama-3B and Llama-8B. Dotted line shows the total number of problems in each respective dataset.}
    \label{fig:stopping_reasons}
\end{figure} 

The termination breakdown (Figure~\ref{fig:stopping_reasons}) closes the loop. For SSDP on GSM8K, frontier depletion accounts for essentially all terminations across both models and all budgets, while DPTS terminations are dominated by budget exhaustion and early stopping. The diagnostic is clean: DPTS is bound by what it can do with the tokens, SSDP is bound by what is left to expand on. Increasing the allocation therefore translates into additional search effort for DPTS but is wasted on SSDP. For flexible infrastructures, where opportunistic compute (a bursting cloud allocation, an idle GPU partition) should turn into additional progress, this is the elasticity test that SSDP fails: it cannot convert headroom into accuracy regardless of the regime.

\subsection{Two Complementary Limitations and the Need for Flexible Search}

Taken together, the two methods define a fundamental tension in ToT reasoning. DPTS struggles at small budgets because it needs time to warm up. Once it has enough data to guide search, it scales well and keeps improving. On the contrary, SSDP gets to a first solution quickly and efficiently, but once the frontier is gone, it cannot do anything more regardless of how much budget remains. More concretely, DPTS is good at finding better solutions once it already has one, but getting that first solution is expensive. SSDP is good at finding a first solution cheaply, but it has no mechanism for improving beyond that. Neither method adjusts its behavior depending on where it is in the search or how much budget is left.

What this points to is that a rigid search strategy, which does the same thing regardless of budget or progress, will always leave something on the table. A more flexible approach, one that can shift its priorities based on what has been found so far and what resources remain, would be better positioned to handle different problem difficulties and compute constraints. The specific form such a strategy should take is an open question, but the gap left by both DPTS and SSDP makes clear that it is worth pursuing.

\section{Conclusions}
\label{sec:conclusions}
This paper presented a controlled side-by-side diagnosis of two representative Tree-of-Thought reasoning strategies: DPTS (MCTS-based) and SSDP (semantic-pruning). We evaluated these ToT strategies across two mathematical reasoning benchmarks, two Llama model scales, and four token budgets. By instrumenting candidate reachability, termination causes, and the allocation-vs-utilization gap, we exposed two limitations that pull in opposite directions: DPTS suffers a structural cold-start bottleneck that makes it unreliable precisely when budgets are tightest, while SSDP collapses through frontier depletion that no amount of additional compute can recover from. Because both methods commit to fixed policies at design time, neither converts the elastic, heterogeneous compute constraints typical of flexible scientific infrastructures into proportional accuracy gains. While our findings are drawn empirically from a small set of benchmarks, we anticipate the observations to be applicable in general to other benchmarks rather than as worst-case bounds for the broader MCTS-based and semantic-pruning families. We leave a broader study that involves more
benchmarks as future work.

The opposing nature of the two limitations also frames another clear future research direction: adaptive ToT strategies that dynamically adjust their exploration at inference time based on what has been observed in the search so far and what compute remains, rather than committing to either exploration-dominated or pruning-dominated behavior up front. Concretely, this calls for budget-aware schedulers that begin in a pruning-friendly regime to amortize the MCTS cold-start, then dilate or contract the pruning aggressiveness (e.g., SSDP's similarity threshold $\tau$) as a function of remaining budget, search progress, model scale, and observed embedding geometry. The metrics introduced in this paper (candidate reachability, termination cause, and the allocation-vs-utilization gap) are the natural feedback signals for such an adaptive strategy.

\begin{acks}
This work is supported in part by the U.S. Department of Energy (DOE),
Office of Science, Office of Advanced Scientific Computing Research
under contract DEAC02-06CH11357/0F-60169 and the National Science
Foundation (NSF) under award no.\ 2411386/2411387, 2514056. AI tools were used in limited capacity to improve clarity, and was fully verified by the authors.
\end{acks}

\balance
\bibliographystyle{ACM-Reference-Format}
\bibliography{main}

\end{document}